\documentclass[10pt, a4paper]{article}
\usepackage{lrec}
\usepackage{graphicx}
\usepackage{tabularx}
\usepackage{soul}

\usepackage{epstopdf}
\usepackage[latin1]{inputenc}

\usepackage{hyperref}
\usepackage{xstring}
\usepackage{svg}
\usepackage{todonotes}

\title{Creating a Multimodal Dataset of Images and Text\\ to Study Abusive Language}

\name{Alessio Palmero Aprosio, Stefano Menini, Sara Tonelli}

\address{Fondazione Bruno Kessler \\
         Via Sommarive 18, Trento (Italy) \\
         \{aprosio, menini, satonelli\}@fbk.eu\\}

\abstract{
In order to study online hate speech, the availability of datasets containing the linguistic phenomena of interest are of crucial importance. However, when it comes to specific target groups, for example teenagers, collecting such data may be problematic due to issues with consent and privacy restrictions. Furthermore, while text-only datasets of this kind have been widely used, limitations set by image-based social media platforms like Instagram make it difficult for researchers to experiment with multimodal hate speech data. We therefore developed CREENDER, an annotation tool that has been used in school classes to create a multimodal dataset of images and abusive comments, which we make freely available under Apache 2.0 license. The corpus, with Italian comments, has been analysed from different perspectives, to investigate whether the subject of the images plays a role in triggering a comment. We find that users judge the same images in different ways, although the presence of a person in the picture increases the probability to get an offensive comment.\\ \newline  }

\begin{document}

\maketitleabstract

\section{Introduction}
The problem of abusive language has recently become very relevant on social media platforms, since it  can quickly escalate and lead to negative outcomes such as cyberbullying, hate speech, and scapegoating. 
To contrast this phenomenon with automated techniques, the NLP research community has therefore started a series of initiatives around workshops \cite{W17-3000,W18-5100,ws-2019-abusive} and shared tasks to detect abusive and hateful speech first on English \cite{Waseem2016HatefulSO,golbeck2017large,DBLP:conf/icwsm/DavidsonWMW17} and more recently also on other languages including German, Italian and Spanish \cite{struss2019overview,HaSpeeDe2018Evalita,basile-etal-2019-semeval}. This has led to the creation of several manually annotated benchmarks in multiple languages, which however mainly rely on platforms that give an easy access to their data, such as Twitter, Wikipedia or Reddit. 

While these platforms are very popular among adults, however, recent surveys show that teenagers and young adults, who are usually more vulnerable when it comes to online attacks, tend to prefer other types of platforms combining both videos, images and texts. For example, a recent study by the Pew Research Center\footnote{\url{http://bit.ly/pewressocial}} showed that SnapChat and Instagram are most popular among 18-to-24 year-olds. This suggests that, in order to contrast hate messages on these platforms, NLP techniques would not be enough, while hate classification tools supporting multi-modality would be more appropriate. In the light of this finding, we introduce in the present paper a tool to create datasets in a simulated scenario, where images are commented in an Instagram-like setting if annotators think that the given image may trigger an offensive comment. The tool, called CREENDER, is web-based and includes a user-friendly interface to write comments to images downloaded from the Web, and to associate a semantic label to each comment, aimed at categorizing the target of the offense. We not only describe the tool in detail, but also the data collection carried out in several school classes. There, the tool was used to write offensive comments related to images, as well as to raise awareness on images and comments posted online. We also present some statistics on the collected dataset, which we release on Github,\footnote{\url{https://github.com/dhfbk/creep-image-dataset}} and perform an analysis of the most common semantic categories assigned during annotation. Finally, we investigate whether there is a correlation between the subject of a picture (male, female, mixed group, etc.) and the offensive message associated with  it. Since the data were collected during a simulation in classes, for which both the parents and the teachers had given a written consent, this dataset can be freely used for research purposes, and does not have the ethical implications that would derive from using real data posted by teenage users on existing social networks.




\begin{figure*}[t]
\centering
\includegraphics[width=8.5cm]{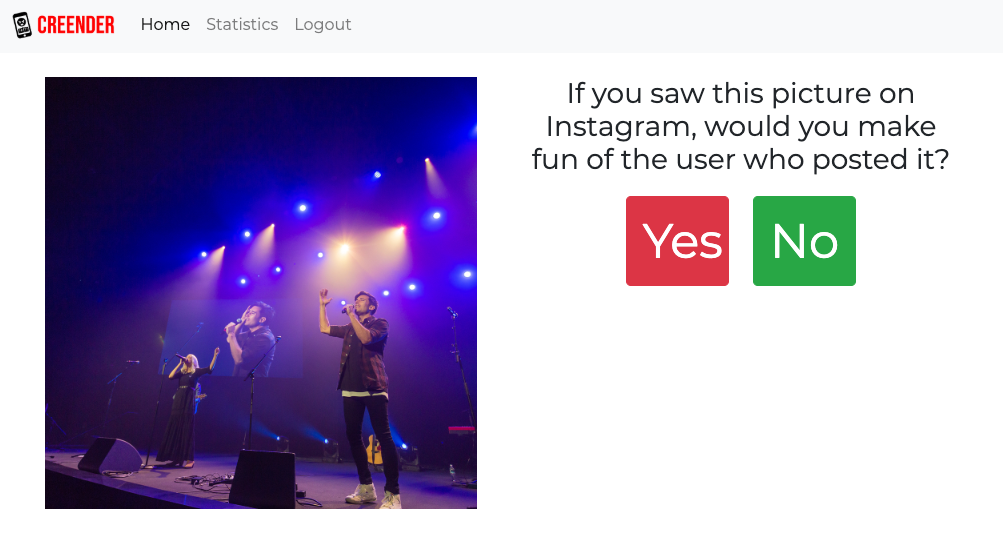}
\includegraphics[width=8.5cm]{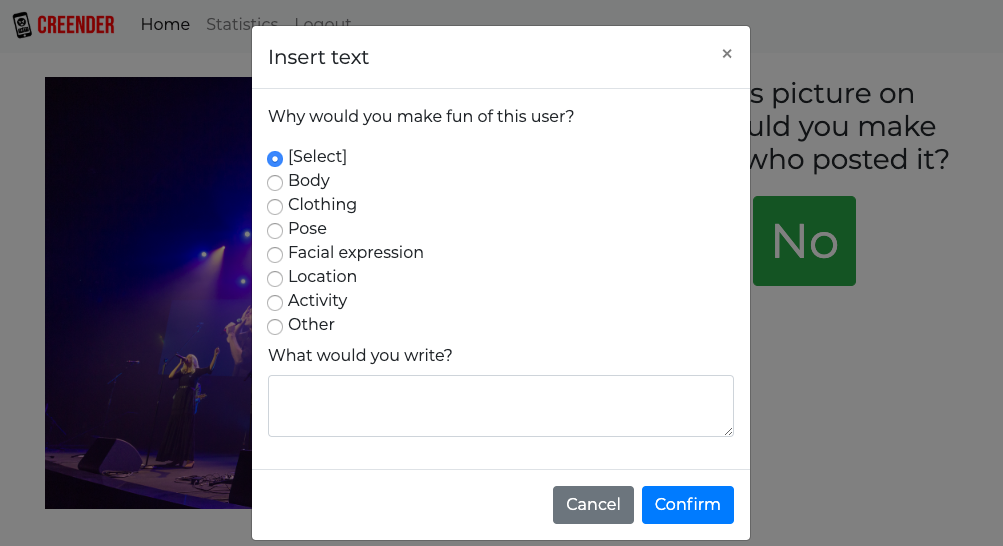}
\caption{CREENDER interface}
\label{fig:screen}
\end{figure*}

\section{Related Work}

Several datasets have been created to study hate speech, abusive language and cyberbullying. Most of them include single textual comments or threads annotated as being hateful/offensive/abusive or not. 
For example \newcite{reynolds2011using} propose a dataset of questions and answers from Formspring.me, a website with a high amount of cyberbullying content. It consists of 12,851 posts annotated for the presence of cyberbullying and severity.  Another resource developed by \newcite{bayzick2011detecting} consists of conversation transcripts (thread-style) extracted from MySpace.com. The conversations are divided into groups of 10 posts and annotated for presence and typology of cyberbullying (e.g. denigration or harassment, flaming, trolling).  For an overview on existing annotation schemes and datasets specific to cyberbullying see the survey presented in \cite{emmery2019current}.
Similarly, a project called Hate Speech Datasets\footnote{\url{https://github.com/leondz/hatespeechdata}} collects a comprehensive list of datasets that are annotated with offensive language, online abuse, and so on.

Probably the most popular datasets shared within the NLP community have been extracted from Twitter because of its relatively easy-to-use APIs. Indeed, most of the shared tasks recently organised to build and evaluate hate speech detection systems use Twitter data \cite{basile-etal-2019-semeval,struss2019overview,DBLP:conf/evalita/BoscoDPST18,aragon2019overview}.

The relationship between textual content and images and the role that they play together is a relatively understudied problem in relation to online hate speech. Few datasets have been created, also because of the restrictions posed by the use of images. A notable exception is the the dataset collected from Instagram  by \newcite{hosseinmardi2015prediction}, which consists of 2,218 media sessions (groups of 15+ messages associated to a media such as a video or a photo), with a single annotation for each media session indicating the presence of cyberagressive behavior. In this dataset the annotation refers to a thread and not to  single offensive messages. 
This dataset has been used for supervised classification tasks, for example in \cite{cheng2019hierarchical} it was employed to detect cyberbullying through a hierarchical attention network that takes into account the hierarchical structure of social media sessions and the temporal dynamics of cyberbullying. Other Instagram datasets have been created but they cannot be shared due to the restrictions in the social network policy \cite{yang-etal-2019-exploring}.

With this work, we address the need to create datasets that on the one hand are compliant with privacy issues and on the other hand are user-friendly and allow the simulation of a real scenario with selected users. The released annotation tool, which can be employed with user-specific images taken from any freely available dataset, is meant to overcome the privacy issues that have hindered so far the creation and sharing of hate speech datasets including text and images.

\section{Annotation Tool}

In order to collect a dataset combining images and offensive comments in a user-friendly way, a web-based system was developed, called CREENDER. This can be accessed both via browser and mobile phone, so that students could use it even if Internet was not available in classes.
The software is distributed as an open source package\footnote{\url{https://github.com/dhfbk/creender}} and is released under the Apache license (version 2.0). It is written in php and relies on a MySQL database. The web interface is multi-language (English, French and Italian already included, other language files can be added as needed), and the language can be assigned at user level, meaning that the interface for users on the same instance can be configured in different languages.



On the configuration side, a set of photos (or a set of external links to images on the web) are given to the tool. Then, one can set the number of users and the number of annotations that are required for each photo. Finally, the system assigns the photos to the users and creates the login information for them.
A single installation of the tool can support multiple instances of annotations.
CREENDER provides a configuration script where the administrator can set the above-described parameters to create the environment.

After a user logs in the system, a picture randomly taken from the images folder is displayed, and a prompt asks ``\textit{If you saw this picture on Instagram, would you make fun of the user who posted it?}''. If the user selects ``\textit{No}'', then the system picks another image randomly and the same question is asked. If the user clicks on ``\textit{Yes}'', a second screen opens where the user is asked to specify the reason why the image would trigger such reaction by selecting one of the following categories: ``\textit{Body}'', ``\textit{Clothing}'', ``\textit{Pose}'', ``\textit{Facial expression}'', ``\textit{Location}'', ``\textit{Activity}'' and ``\textit{Other}''. The user should also write the textual comment s/he would post below the picture. After that, the next picture is displayed, and so on.
A screenshot of the tool is displayed in Figure \ref{fig:screen}. 


\begin{table*}[h]
    \centering
    \begin{tabular}{lrrr}
    \hline \hline
     Pictures with $\downarrow$ ... and having $\rightarrow$ & At least 1 comment   & (Total comments) & No comments\\
        \hline
        At least 1 judgement    &    1,018 & 1,135 & 16,894 \\
        At least 2 judgements    &    901 & 1,018 & 9,876 \\
        At least 3 judgements    &  713 & 815 & 5,454 \\
        At least 4 judgements    &  495 & 563 &  3,060\\
    \hline \hline     
    \end{tabular}
    \caption{Number of pictures in the dataset with at least \textit{n} judgments (\textit{`yes'/`no'}) and number of comments.}
    \label{tab:annotationNumbers}
\end{table*}

The question posed by the system does not ask explicitly whether the user would \textit{insult}, \textit{harrass} or \textit{offend} the person who posted the image, because in a preliminary test with students we observed that the answer would almost always be ``\textit{No}''. This showed that only in few cases a user would consciously harm another user, especially if the two know each other.  Furthermore, comments with explicit hateful content are easy to find online and can be unambiguously annotated in most of the cases. We therefore decided to focus on a more nuanced form of offensive message, that is when a user makes fun of another one. We made this choice because we assumed that this kind of messages would be more ambiguous, containing ironic or sarcastic comments, and mixing humorous and abusive content without being necessarily explicit. This would make the collected data very interesting from a linguistic and computational point of view. However, the prompt can be easily modified, asking users for other types of comment.

\section{Annotation Process}
\label{annotproc}
The CREENDER tool was used to collect abusive comments associated to images, simulating a setting like Instagram in which pictures and text together build an  interaction which may become offensive. The data collection was carried out in several classes of Italian teenagers aged between 15 and 18, in the framework of a collaboration with schools aimed at increasing awareness on social media and cyberbullying phenomena. 
The data collection was embedded in a larger process that required two to three meetings with each class, one per week, involving every time two social scientists, two computational linguists and at least two teachers. During these meetings several activities were carried out with students, including simulating a WhatsApp conversation around a given plot as described in \cite{W18-5107}, commenting on existing social media posts, and annotating images as described in this paper. Since ethical issues were a main concern since the drafting of the study design, because all participants were underage students, all the activities had been co-designed with the schools involved and informed consent was gathered beforehand both from teachers and from parents. 

Overall, 95 students were involved in the annotation. The sessions were organised so that different school classes annotated the same set of images, in order to collect multiple annotations on the same pictures. However, since some users were quicker than others in giving a judgement on the pictures, we could not collect multiple annotations for all images included in the dataset (see Table \ref{tab:annotationNumbers} for details).

\section{Annotated Corpus}

Overall, 17,912 images have been judged at least once by the students. For 1,018 of them, at least an offensive comment has been written during the annotation sessions. Overall, the number of comments in the dataset is 1,135, which is higher than the number of pictures with a comment because the same image may be commented more than once by different students. An overview of the content of the dataset including images and comments is presented in Table \ref{tab:annotationNumbers}. Note that the number of  \textit{judgements} refers to the `yes/no' option selected by users in the first platform view,  while the number of \textit{comments} refers only to the images tagged with a `yes', for which a student wrote also a comment.

\begin{figure}[h]
\centering
\scalebox{0.3}{
  \includegraphics{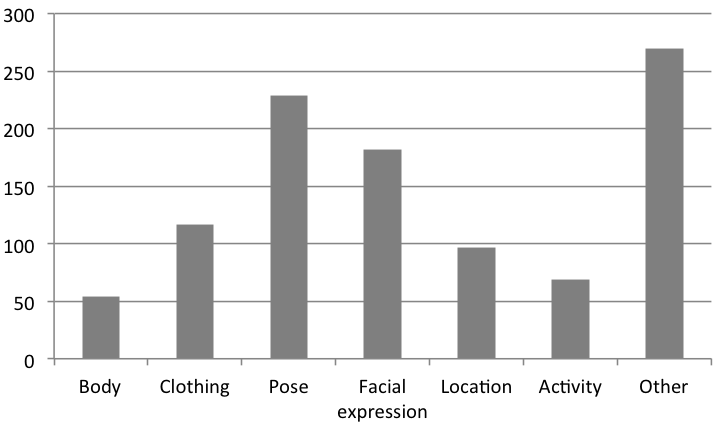}
 }
  \caption{Distribution of trigger categories assigned to the  comments}\label{label-comments}
\end{figure}

Overall, only one image has been tagged with four `yes', and in most of the cases annotators selected only `no'. The number of images tagged with exactly three `yes' is 13, those with two is 88.
Since these images have been leveraged from Instagram with no particular criterion in mind, the distribution of `yes' and `no' may be considered realistic, with the majority of pictures not triggering any potentially negative reaction, and around 6\% of them being associated with offensive comments.

In Figure \ref{label-comments} we report the distribution of \textit{trigger categories} associated with the comments, i.e. the reason why each annotator would make fun of the user who posted the image. The categories are rather imbalanced, with a minority of comments associated with the `Body' label and  several comments concerning the `Pose'. However, for most of the comments the `Other' label was used. When collecting feedback from students after the annotation sessions, several annotators suggested that it should be possible to assign multiple labels instead of just one, and reported that they used the `Other' label for those cases. On the other hand, they did not express the need to include additional categories in the annotation task.

In general, we observe that there is a low agreement on whether a picture triggers an offensive comment or not. This suggests that an offensive intent is more dependent on the attitude of a user posting a comment than on image-specific features. 
We also compute inter-annotator agreement -- using Krippendorff's alpha measure \cite{krippendorff1970estimating} -- on the trigger categories assigned to the comments, considering only the images that received at least two comments. Agreement is 0.19, which implies again that the reason to make fun of a user does not depend on a specific feature of the picture, but rather that multiple aspects of a posted image can be taken as an excuse for offensive comments. In order to avoid the ambiguity introduced by the \textit{`Other'} label, we also compute IAA 
ignoring this class. This time the agreement value is 0.26, showing that on the one hand the `Other' label covers uncertain cases, but also that the reason to comment a picture remains highly subjective.

As an example, we report below some of the comments to the pictures that were written during the annotation sessions. They confirm that, although we did not explicit mention in the prompt that comments should be offensive, they were interpreted in this way by annotators:

\begin{enumerate}
\item \textit{Copriti (Cover yourself up)}
\item \textit{Che schifo di foto (This picture sucks)}
\item \textit{Inquietante (Disturbing)}
\item \textit{Che gusti orribili, ma cos'\`{e}? (What a bad taste, what's this?)}
\end{enumerate}

The above examples show that these comments have different features compared to hate speech messages extracted from Twitter: they tend to be short because they complement the image and they are rich in deictic expressions. In most of the cases, they are not self-contained from a semantic point of view.

\begin{table}[]
    \centering
    \scalebox{0.9}{
    \begin{tabular}{lrrrr} 
    \hline
    \hline
          & \textbf{Females} & \textbf{Males} & \textbf{Mixed} & \textbf{Nobody} \\
    \hline
        \textbf{Body}              & 27  & 20 & 3 & 4 \\
        \textbf{Clothing}          & 66  & 30 & 9 & 12 \\
        \textbf{Pose}              & 114 & 99 & 11 & 5 \\
        \textbf{Facial Expression} & 68  & 90 & 17 & 7 \\
        \textbf{Location}          & 16  & 17 & 7 & 57 \\
        \textbf{Activity}          & 12  & 14 & 7 & 36 \\
        \textbf{Other}             & 72  & 63 & 22 & 113 \\
        \hline
         \textbf{Total}             & 377  & 318 & 76 & 252 \\
         \hline
    \hline
    \end{tabular}}
    \caption{Distribution of the triggers over the subject types.}
    \label{tab:distributionlab}
\end{table}

\begin{figure*}[h]
\centering
\scalebox{0.5}{
  \includegraphics[]{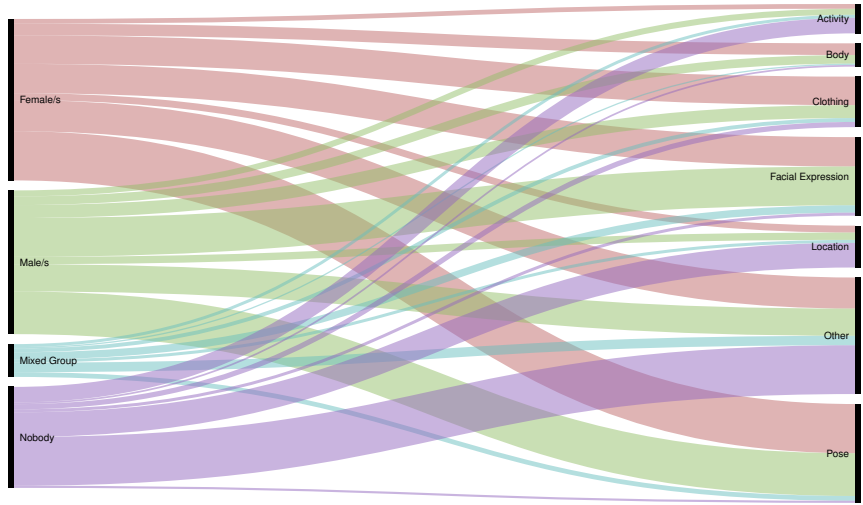}
 }
  \caption{Analysis of pictures with at least one annotation: the subject types (left) are put in connection with the trigger categories  (right)}\label{alluv-categ}
\end{figure*}


\section{Corpus Analysis}

In order to analyse whether the subject of a picture has an impact on the choice to write an offensive message, we perform two different analyses. We first assign manually  the subject of each image having at least 1 comment to one of the following categories: male-only subject(s), female-only subject(s), mixed group, no human subject. We show in Figure \ref{alluv-categ} how the different subject types relate to the trigger categories. The graphical representation shows that the main differences between pictures with male and female subjects concern the `Facial expression' and `Clothing' categories: the first is more frequently associated with male subjects, while the second seems to be more related to female subjects. When there are multiple subjects with different genders, instead, no particular differences are observed. As expected, when no person is portrayed in the picture, `Location', `Activity' and `Other' are prevalent. In some cases, `Pose' or `Expression' are selected, because of the presence of animals or drawings in the images. The number of annotations for each category is reported in Table \ref{tab:distributionlab}.

\begin{table}[h]
    \centering
    \begin{tabular}{lrr}
\hline \hline
    		&		\% Yes	&	\% No \\ \hline
Females	&	36.85	&	32.14 \\
Males	&	31.09	&	19.00 \\
Mixed	&	7.43	&	9.33 \\
Nobody	&	24.63	&	39.53 \\
\hline \hline
    \end{tabular}
    \caption{Subject types for pictures annotated with `Yes' (i.e. triggering a comment) and `No'}
    \label{tab:distribution}
\end{table}

We perform the same manual classification on a set of 3,200 pictures randomly taken from the images that were tagged with `No', assigning each of them to one of the four subject categories. Then we compare the two distributions, that are reported in Table  \ref{tab:distribution}, by applying the $\chi^2$ test ($N=4,218$), showing a statistically significant difference between the two distributions of categorial variables ($p<.001$). In particular, pictures with no human subject are less likely to get an offensive comment, while those with a female subject are the most commented ones. Also male subjects, however, trigger  offensive comments very frequently, while they are only present in 19\% of the images which were not commented by users.  

\section{Release}
Due to the policies from Instagram and the restrictions on distributing contents from their platform it is not possible to share the original annotated images. Following the strategy adopted in \cite{kruk2019integrating}, for each image we release the last layer obtained with ResNet-18 \cite{he2016deep}, a convolutional neural network pre-trained on the ImageNet database \cite{ImageNet_VSS09} with more than a million images.
Each image representation is provided with all the annotations collected i.e. \textit{yes/no} judgements (with the related comments if the judgment is \textit{yes}), the trigger category (e.g. \textit{facial expression}, \textit{pose}) and the category of the subject/s (\textit{female}, \textit{male}, \textit{mixed} or \textit{nobody}).
The number of annotations for each picture may vary between 1 to 4.

\section{Discussion}
\label{sec:discuss}
Organising annotation sessions in school classes to create the corpus described in the previous sections raises a number of issues. The main concern when creating this kind of datasets is the ethical one, especially in our case involving teenagers. A simulated setting was indeed implemented as it did not require  gathering sensitive information regarding minors, since we assigned to each user a login and a password but did not associate them to their real identity. No information concerning lived experiences was collected. Also, the 20,000 pictures uploaded in the tool and available for annotation had all been briefly checked by researchers to avoid showing pornographic or other kinds of potentially harmful images.

Beside these formal considerations, ethics has been a central concern of the research team throughout the data creation process and it was addressed by-design as far as possible. As briefly mentioned in  Section \ref{annotproc}, the  activity was part of a larger set up. The purpose was to ensure a framing of the experimentation and its possible outcomes in a broader perspective, allowing students to understand the role of the bully and of the victim when offensive messages are posted online. Therefore, the annotation sessions were usually introduced by a lecture in class addressing the issue of cyberbullying, followed by a role-playing game using WhatsApp chats. This represented a protected space to experiment cyberbullying and playing different roles in the interactions. Next, the annotation of this dataset using the CREENDER tool exposed students to real images available online, giving them the possibility to reflect on the content of the pictures and on the reasons why abusive comments are written. Finally, the participatory analysis of the outcome of the annotation during the last session allowed students to exchange ideas, among others, on the low agreement scores, on the content of their comments, and on the subject of the most criticised pictures.

An open issue is the validity of the corpus created with the experimentation. Can the corpus be considered a substitute for actual comments to images that can be found online, or should it be seen just as the result of an experimentation with little (if any) resemblance to reality? Since only few datasets are available to study the use of hate speech on Instagram, and none of them is in Italian, it is difficult to perform a comparison to assess the plausibility of the messages collected during the annotation sessions. A manual check of the comments showed that they include all the offense types that are usually listed under `hate speech': profanities, curses, racist messages, homophobic ones, sexism, sarcastic comments, etc. This suggests that participants avoided self-censorship thanks to the anonymity granted during the annotation. While we are aware that only a thorough methodology could respond to the question of the validity of the corpus gathered through the platform, these preliminary findings suggest that participants may act in a way that resembles real life.

\section{Conclusions}

In this work we present a methodology and a tool, CREENDER, to create multimodal datasets in the hate speech domain. In this framework, participants in online annotation sessions judge whether images may trigger an offensive comment, leave a possible comment and also assign to it a trigger category. The tool is freely available with an interface in three languages, and allows setting up easily annotation sessions with multiple users.

The tool was developed in the framework of activities with schools around the topic of cyberbullying, involving 95 Italian high-school students. CREENDER was therefore used to raise awareness on the images posted online and on the offensive content of some comments, while supporting the creation of a dataset to study these phenomena.  

The analysis of the collected data gives interesting insights into how Instagram-like platforms work. First of all, images containing persons are more likely to trigger offensive comments than those without a human subject. Both female and male subjects are offended but the reasons may differ: the former are offended more because of the pose and of the clothing, while the latter for the pose and the facial expression. In general, the reasons why a comment is triggered seem to be subjective, depending on the user leaving the comment rather on some actual characteristics of the person portrayed in the picture.

We conducted our data collection using Instagram pictures randomly taken from this platform, because it is the social network that is most used by teenagers, including those involved in our annotation sessions. However, this makes the release of the full dataset impossible, sice Instagram has a very restrictive policy concerning images. We therefore adopt a strategy already used within the NLP research community, releasing the images as a layer of a ResNet-18 neural network trained on ImageNet. The comments, instead, are freely available without restrictions due to the consent signed by all parents and by the anonymity granted to participants. This represents a very interesting dataset from a research point of view, since it includes comments written by underage students that are usually difficult to obtain because of privacy reasons.

In case other researchers are interested in using the tool and replicating the annotation methodology, we suggest that freely available repositories of images are used instead of Instagram pictures, such as the Flickr dataset\footnote{\url{https://bit.ly/flickr-image-dataset}} or ImageNet \cite{ILSVRC15}.

In the future, we plan to extend our study to compare the judgements given by single users to those given by groups of peers. In a preliminary study, we observed that, when students are given the possibility to discuss with a small group of peers whether they would like to write an offensive comment, and they can decide together the content of the message, they tend to be more aggressive and are more likely to select `yes'. While the comments collected so far with groups of annotators are not enough to allow a fair comparison between the two settings (single vs. group), we plan to   extend them in the future and pursue also this interesting research line.



\section{Acknowledgements}

Part of this work was funded by the CREEP project,\footnote{\url{http://creep-project.eu/}} a Digital Wellbeing Activity supported by EIT Digital in 2018. This research was also supported by the HATEMETER project\footnote{\url{http://hatemeter.eu/}} within the EU Rights, Equality and Citizenship Programme 2014-2020. In addition, the authors want to thank all the students and teachers who participated in the experimentation.

\section{Bibliographical References}
\label{main:ref}

\bibliographystyle{lrec}
\bibliography{aaai-2020}


\end{document}